%% file: main.tex
\documentclass[10pt,twocolumn,letterpaper]{article}

\usepackage{siunitx}
\usepackage{amsmath}
\usepackage{fontawesome5}
\usepackage{soul}
\usepackage{multirow} 
\usepackage[dvipsnames]{xcolor}

\usepackage[pagenumbers]{cvpr} 

\definecolor{cvprblue}{rgb}{0.21,0.49,0.74}
\usepackage[pagebackref,breaklinks,colorlinks,citecolor=cvprblue]{hyperref}

\def\paperID{3} 
\def\confName{CVPR}
\def\confYear{2024}

\title{NurtureNet: A Multi-task Video-based Approach for Newborn Anthropometry
}

\author{
Yash Khandelwal$^{1\dagger}$
\hspace{5.5mm}
Mayur Arvind$^1$ \hspace{5.5mm}
Sriram Kumar$^1$ \hspace{5.5mm}
Ashish Gupta$^1$ \\
Sachin Kumar Danisetty$^1$ \hspace{5.5mm}
Piyush Bagad$^2$ \hspace{5.5mm}
Anish Madan$^2$ \hspace{5.5mm}
Mayank Lunayach$^2$ \\
Aditya Annavajjala$^2$ \hspace{5.5mm}
Abhishek Maiti$^2$ \hspace{5.5mm}
Sansiddh Jain$^2$ \hspace{5.5mm}
Aman Dalmia$^2$ \\ 
Namrata Deka$^2$ \hspace{5.5mm}
Jerome White$^2$ \hspace{5.5mm}
Jigar Doshi$^2$ \hspace{5.5mm}
Angjoo Kanazawa$^3$ \\
Rahul Panicker$^2$ \hspace{5.5mm}
Alpan Raval$^1$ \hspace{5.5mm}
Srinivas Rana$^1$ \hspace{5.5mm}
Makarand Tapaswi$^{1\dagger}$
\vspace{2mm} \\
Wadhwani Institute for Artificial Intelligence (WIAI) \\
{\small
$^1$currently at WIAI, $^2$work done while at WIAI; $^3$UC Berkeley } \\
{\small 
$^\dagger$ \texttt{\{yash,makarand\}@wadhwaniai.org}}
\vspace{-2mm}
}

\newcommand{\model}{NurtureNet}
\newcommand{\mcU}{\mathcal{U}}
\newcommand{\mcV}{\mathcal{V}}
\newcommand{\bx}{\mathbf{x}}
\newcommand{\bz}{\mathbf{z}}
\newcommand{\MLP}{\text{MLP}}
\newcommand{\wgt}{w_\text{gt}}
\newcommand{\mySIrange}[3]{\qtyrange[range-units = single]{#1}{#2}{#3}}

\renewcommand{\paragraph}[1]{\vspace{1mm}\noindent\textbf{#1}}

\begin{document}
\maketitle

\input{sec/0_abstract}
\input{sec/1_intro}
\input{sec/2_relwork}

\input{sec/3_method}
\input{sec/4_experiment}

\input{sec/5_conclusion}

{
\small
\bibliographystyle{ieeenat_fullname}
\bibliography{longstrings,refs,relwork_refs}
}

\clearpage
\input{sec/6_suppl}

\end{document}

%% file: sec/0_abstract.tex
\begin{abstract}

Malnutrition among newborns is a top public health concern in developing countries.
Identification and subsequent growth monitoring are key to successful interventions.
However, this is challenging in rural communities where health systems tend to be inaccessible and under-equipped, with poor adherence to protocol.
Our goal is to equip health workers and public health systems with a solution for contactless newborn anthropometry in the community.

We propose \model{}, a multi-task model that fuses visual information (a video taken with a low-cost smartphone) with tabular inputs to regress multiple anthropometry estimates including weight, length, head circumference, and chest circumference.
We show that visual proxy tasks of segmentation and keypoint prediction further improve performance.
We establish the efficacy of the model through several experiments and achieve a relative error of 3.9\% and mean absolute error of \SI{114.3}{\gram} for weight estimation.
Model compression to \SI{15}{\mega\byte} also allows offline deployment to low-cost smartphones.

\end{abstract}

%% file: sec/1_intro.tex
\vspace{-2mm}
\section{Introduction}

The first 4 weeks of life are critical for a newborn’s physiological and neurological development.
Conditions such as malnutrition and malabsorption during this phase lead to neonatal morbidities and in extreme cases even mortality.
Thus, tracking a newborn's growth over the first few weeks is an important public health responsibility~\cite{world2009child}.

The weight of a newborn is an important statistic that captures its overall health and well-being~\cite{christian2013risk, gu2017gradient, jornayvaz2016low, blencowe2019national}.
Other measurements such as length, head circumference, and chest circumference are also useful for assessing growth or related developmental disorders~\cite{microcephaly,lbw1985}.
However, there are several challenges in accurately capturing it in low- and middle-income countries (LMICs).

\begin{figure}[t]
\centering
\includegraphics[width=0.77\linewidth]{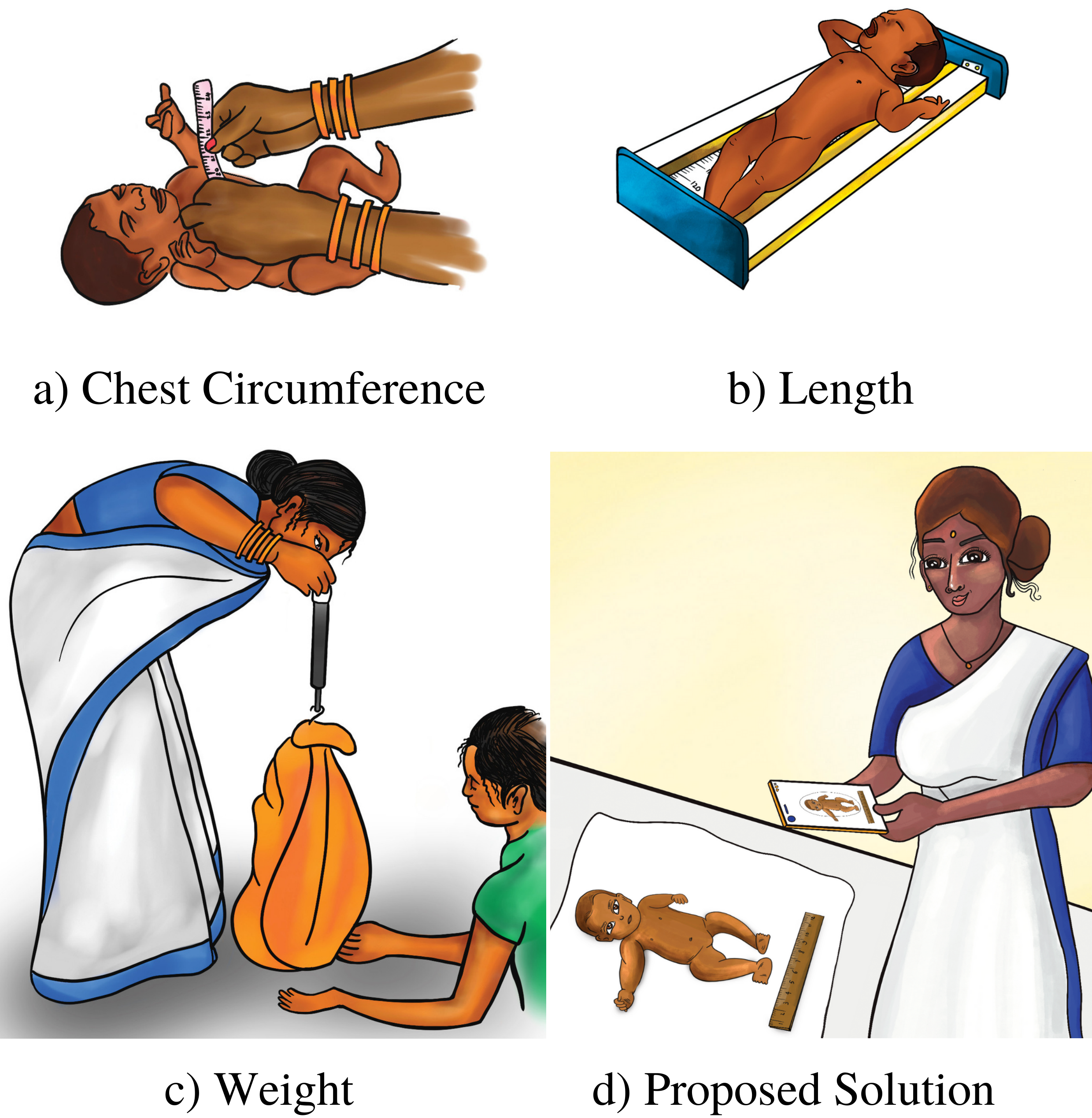}
\caption{Illustration contrasting traditional approaches (a-c) for newborn anthropometry to what our proposed solution (d) enables.
(a)~A measuring tape is used to measure head and chest circumference.
(b)~An infantometer is used to capture length.
(c)~The newborn is suspended from a cloth and hooked up to a spring balance to measure weight.
(d) Our proposed solution replaces all the above tasks and only requires the data collector to take a short video with a low-cost smartphone.
}
\vspace{-3mm}
\label{fig:teaser}
\end{figure}

As seen in Fig.~\ref{fig:teaser}(c), traditional methods for measuring weight in community settings use a spring balance (least count \SI{100}{\gram}), from which the newborn is suspended.
This results in two main sources of error:
(i)~Human factors:
the panicked mother supporting her baby from the bottom;
motion of the spring balance as the newborn moves;
difficulty in ascertaining the reading due to parallax;
cultural challenges such as reluctance towards ``outsiders" handling babies; and
data handling malpractices leading to reporting challenges.
(ii)~Instrument factors: old machines whose springs are no longer taut result in positive errors (over-prediction);
poorly calibrated or uncertified instruments; and even
unavailability of the instrument due to supply chain issues.
Similar challenges also apply to other anthropometric measurements such as newborn length or head and chest circumference.

There are also logistical factors at play.
Rural communities may be several miles away from health facilities with poor mobility options and limited inter-connectivity. 
Poverty further affects their ability to avail health facilities.
Geographical barriers like rough terrain or rivers, and seasonal challenges such as extreme heat and heavy rain make it challenging for both, families (with newborns) to reach health centers and for health workers (carrying heavy instruments) to visit rural communities.

Our goal is to develop a contactless, geo-tagged, easy-to-use solution that provides accurate anthropometry estimates for a newborn (age 0-42 days).
We wish to leverage the proliferation of mobile phone adoption in rural areas of LMICs enabling AI technologies to improve the daily activities of frontline health workers while ensuring automatic reporting for timely public health response and policy formulation.
Our technology is suitable primarily in rural community settings over health facilities that may have good instrumentation and well-trained staff.
In such rural settings, there are about 1 million health workers in our country, making this solution amenable for large-scale impact.

To facilitate widespread adoption, we make several design choices.
(i)~We restrict ourselves to RGB videos captured on low-cost mobile devices and forego complex depth sensors that may curtail adoption in rural areas.
(ii)~A reference object is needed to provide a sense of metric scale and we use easily available
wooden rulers
instead of chessboards.
(iii)~We develop a simple protocol for capturing the video that enables viewing the newborn from multiple angles without the need for a dedicated video capture setup or specialized hardware.
(iv)~We restrict modeling to simple architectures that can be 
compressed and deployed on a low-cost smartphone to enable offline inference.

We present a CNN-based model that can ingest video frames, aggregate their information, augment it with relevant tabular inputs, and estimate the newborn's weight.
This weight estimation model is extended to estimate several anthropometric measurements through a multi-task setup (\model{}).
We also propose proxy tasks such as baby segmentation and keypoint estimation that can assist the model in focusing on the baby's shape and pose respectively, resulting in improved performance.
Notably, building on state-of-the-art segmentation and keypoint estimation methods, we show that segmentation masks and keypoints need not be annotated for each frame, and pseudo-labels can be used instead.

Overall, our contributions can be summarized as follows:
(i)~We present a vision application for newborn anthropometry based on a standard RGB video captured with a low-cost smartphone that enables widespread adoption and impact in rural community settings.
(ii)~We propose and train a multi-task model, \model{}, that ingests the video and is assisted by tabular inputs to estimate several anthropometric measurements simultaneously.
(iii)~We show the benefits of modeling auxiliary vision tasks (\eg~segmentation and keypoints) with pseudo-labels for training anthropometry models; and
(iv)~We present thorough experiments to show the impact of various modeling ideas, including evaluation of compressed models for offline phone deployment.

%% file: sec/2_relwork.tex
\section{Related Work}

Vision techniques have been used for various applications in newborn and infant healthcare, specially for anthropometry.

\paragraph{Computer vision for newborns}
is used in applications such as heart rate monitoring~\cite{scalise_2012heart},
General Movement Assessment (GMA) for early detection of cerebral palsy~\cite{meinecke2006movement, schroeder2020gma},
postnatal age estimation~\cite{torres2019postnatal},
and even identification based on footprints~\cite{liu2017infant}.
Related to anthropometry,
there is work on estimating birth weight using ultrasound videos prior to birth~\cite{plotka_2022babynet}.
An infant's length is estimated using easy-to-detect stickers or markers~\cite{tang_2018measuring},
children's (aged 2-5 years) height based on point clouds~\cite{trivedi2021height},
and even height for adults using multiple images and a large reference object~\cite{liu2018single}.
However, the above methods require specialized hardware or equipment and are therefore not applicable in low resource areas.
A different approach, Baby Naapp~\cite{fletcher_2017development} aims to use vision tools to eliminate the need for manual transcription by capturing and analyzing images/videos of devices - spring balances for weight and measuring tapes for circumference.
Closest to our work, single image based weight and height estimation is performed using CNN-based regression~\cite{shu2023single}.
However, their goal is different from ours as they aim to clinically estimate birth weight by capturing images in controlled settings (hospital).
We show that such an approach performs poorly in community settings where protocol adherence may be difficult.

\paragraph{Pose estimation}
for newborns or infants is a popular task.
Tracking body movements for newborns is critical, and early detection of abnormalities can prevent long-term health effects~\cite{schroeder2020gma}.
Specifically, tracking baby pose over time is useful to perform GMA~\cite{silva2021gmareview}, indicative of conditions such as cerebral palsy, autism spectrum disorder, and Rett syndrome.
Approaches for infant body pose estimation include handcrafted features such as histograms of 3D joints~\cite{mccay2019establishing} or Random Ferns on depth images~\cite{hesse_2015estimating}.
Depth-only videos have also been used along with CNNs to directly regress pose~\cite{moccia_2019preterm_ST, wu2022_supine}. 
Towards GMA, CNN-based pose regression models have been developed that work with RGB or RGB-D data~\cite{groos_2022towards, abbasi_2022deep, ni_2023semi, kyrollos2022transfer}.
Transformer models are making inroads in infant pose estimation with RGB images~\cite{Cao_2022AggPoseDA} or depth and pressure images~\cite{kyrollos_2023}.
As a proxy to pose estimation, body part segmentation may also be used to understand infant movement~\cite{zhang_2019online}.
3D parametric models and pose estimates are also used to estimate the height and weight of adults~\cite{thakkar2022reliability}.
For anthropometry, we show that segmentation and keypoint detection are good proxy tasks that help the model focus on the newborn.

\paragraph{3D parametric models for adults and infants.}
The Skinned Multi-Person Linear (SMPL)~\cite{loper2015smpl} model has been wildly popular in modeling the 3D shape of a human and
has stemmed a flurry of methods~\cite{hmr,meshtransformer,cmr,probabilisticmesh,pose2mesh,pixel2mesh,coherent3dpose}.
However, adult human models cannot be used directly for newborns, predominantly due to changes in body shape proportions~\cite{huang_2021invariant, sciortino2017estimation}.
Hence, a Skinned Multi-Infant Linear (SMIL) model was proposed~\cite{hesse_2018learning}.
Unfortunately, the model does not fit well to our case as it is trained on European infants in the 2-4 months age range with a significantly higher weight distribution.
In contrast, we are interested in anthropometry for newborns up to 42 days of age in LMICs.

\paragraph{Tabular methods.}
Contrary to vision-based approaches, tabular data in the form of electronic health records (EHR) are used for infant~\cite{khan2022infant} and fetal weight~\cite{lu_2019ensemble} prediction.
These methods use maternal attributes, economic factors, and other aspects related to the gestation period as predictive features~\cite{kader2014socio, senthilkumar2015prediction, kuhle2018comparison, loreto2019predicting}.
However, it can be hard to train or deploy models in regions where such records are inaccessible or not carefully curated.
In our work, we use tabular data (birth weight and age) to augment and assist visual features.
Importantly, we show that our vision model augmented with tabular inputs is robust to errors in the tabular data that may be common due to misreporting.

%% file: sec/3_method.tex
\section{Method}
\label{sec:method}

We formulate anthropometry as a regression problem and introduce an end-to-end pipeline to estimate the weight~$w$, length~$l$, head circumference~$h$, and chest circumference~$c$ of an infant with age $a \in [0, 42]$ days.
Our model ingests a video $\mcV$ and is augmented by tabular information such as the birth weight $w^0$ and the current age to regress:
\begin{equation}
[w, l, h, c] = f_\theta(\mcV, w^0, a) \, ,
\end{equation}
where $\theta$ are the model's learnable parameters.

\begin{figure*}[t]
\centering
\includegraphics[width=\linewidth]{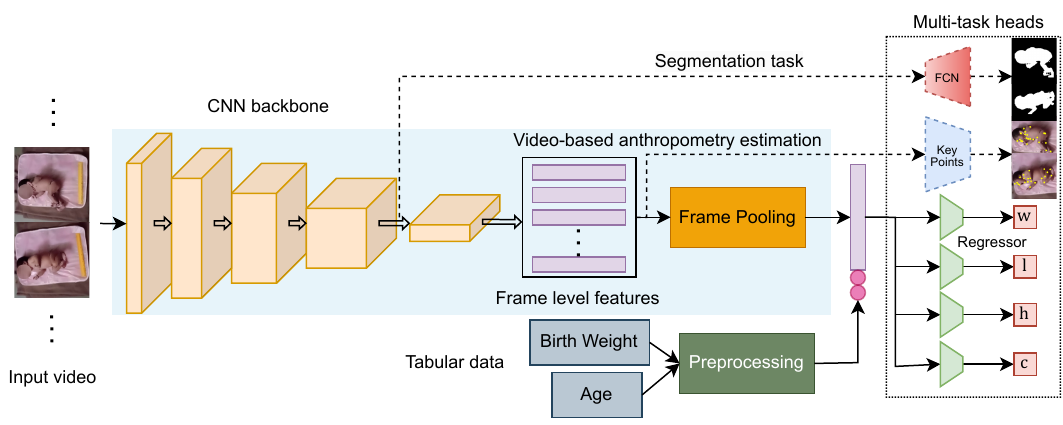}
\vspace{-6mm}
\caption{Overview of the proposed approach.
Input video frames are sub-sampled and processed using a CNN and fused using a pooling module.
Tabular data is normalized between $[0, 1]$ and concatenated to this video representation.
We use independent MLP regressors to predict anthropometry measures: weight, length, head circumference, and chest circumference.
Additionally, we introduce two proxy tasks only used during training: newborn pixel segmentation predicted through an FCN head and keypoint estimation through a simple MLP.}
\vspace{-2mm}
\label{fig:method}
\end{figure*}

The visual component of the model learns an implicit shape representation through the fusion of multiple frames that capture the newborn from several angles (Sec.~\ref{subsec:method:video-based_weight}).
Furthermore, we encourage the model to focus on the newborn by asking it to predict a segmentation mask and keypoints
in a bootstrapped multi-task setting (Sec.~\ref{subsec:method:multitask}).
Finally, we show how the visual features can be augmented with tabular data, resulting in significant improvements (Sec.~\ref{subsec:method:tabular}). Fig.~\ref{fig:method} illustrates the overall approach.

\subsection{Video-based Anthropometry}
\label{subsec:method:video-based_weight}

\paragraph{How to record a video?}
Before addressing modeling, we briefly talk about how we record the video $\mcV$.
Predicting the metric shape of an entity using a monocular camera often requires a reference object.
However, considering the large-scale rural use-case of our solution,
we switch from the chessboard (\faChessBoard{} a classic and accurate reference object used for camera calibration~\cite{szeliski2022computer}) to a wooden ruler (\faRuler{} length \SI{30}{\centi\meter}) that is easily available to health workers.
The newborn is laid on a bedsheet spread on a flat surface with a \faRuler{} placed below the newborn (in the same plane). 
While we remove all clothes for the newborn, no specific instructions are provided for the bedsheet.
The data collectors (or health workers) are trained to capture a video by starting from the top of the baby and making a smooth arc as illustrated in \cref{fig:video_collection}.
We filter videos by quality
to ensure the newborn and reference object are clearly visible for a majority of the video (see~\cref{sec:app:data_validation}).

\paragraph{Video-based weight estimation.}
Consider a video $\mcV = [f_1, \ldots, f_T]$ with $T$ frames.
We sample $N < T$ frames from the video and pass them through a CNN backbone $\phi(\cdot)$ to obtain frame-level representations $\bx_l = \phi(f_l)$ for each selected frame $f_l$, where $\bx_l \in \mathbb{R}^d$.
Our approach combines the individual frame-level representations via a pooling function $\bz = \rho(\{\bx_l\}_{l=1}^N)$, and is followed by a Multi-Layer Perceptron (MLP) with one hidden layer to estimate the weight $w \in \mathbb{R}^1$ in \unit{\kilo\gram}.

We explore several pooling functions $\rho(\cdot)$ ranging from average and max-pooling to complex ones involving vanilla attention~\cite{bahadanau2014attention}.
Interestingly, given the nature of the problem, we find that a permutation invariant pooling (such as max-pooling) provides good results, as suggested by previous works on 3D shape~\cite{su20153dshape}.
We estimate the weight:
\begin{equation}
\label{eq:mlpw}
w = \MLP_w( \max(\bx_1, \ldots, \bx_N) ) = \MLP_w( \bz ) \, .
\end{equation}
Similar to work in action recognition~\cite{zhou2018trn}, during training, we randomly select $N$ frames from the video as a form of data augmentation, while during inference, we pick linearly spaced frames.
The model parameters including the CNN backbone $\phi(\cdot)$ are trained using the L1 loss:
\begin{equation}
\label{eq:loss_weight}
L_w = | w - \wgt | \, ,
\end{equation}
where $\wgt$ is the ground-truth weight of the newborn (in \unit{\kilo\gram}).

\input{tables/anthropometric_correlations}

\subsection{Multi-task Learning}
\label{subsec:method:multitask}

Multi-task learning generally leads to performance improvements when the tasks are related to each other~\cite{standley2020tasks}.

\paragraph{Anthropometric measurements.}
Along with weight, we also predict other measurements such as length, head circumference, and chest circumference.
Naturally, a taller baby is likely to be heavier, or a baby with a bigger chest may be better off with respect to nutrition.
We compute the Pearson correlation coefficients across pairs of anthropometric measurements on our training set.
As seen in Table~\ref{table:anthropometric_correlations}, weight is strongly correlated with length, head circumference, and chest circumference.

We attach additional task heads, similar to the MLP used for weight estimation, to the pooled video representation.
Specifically, we create three new task heads to predict each measurement in \unit{\centi\meter}:
length $l = \MLP_l(\bz)$,
head circumference $h = \MLP_h(\bz)$, and
chest circumference $c = \MLP_c(\bz)$.
The model is trained jointly to optimize:
\begin{equation}
\label{eq:loss_anthro}
L_\text{anthro} = \lambda_w L_w + \lambda_l L_l + \lambda_h L_h + \lambda_c L_c \, ,
\end{equation}
where $L_l, L_h, L_c$ are L1 losses applied to length, head circumference, and chest circumference respectively; and $\lambda_{(\cdot)}$ are loss weight coefficients.

\paragraph{Visual prediction tasks.}
While all the above tasks require ground-truth measurements to be collected at the time of video capture, we now present visual tasks that can be annotated post data collection.
In particular, we consider pixel-level newborn segmentation and keypoint estimation, with the intent to encourage the model to learn representations that focus on the newborn.

While annotating segmentation masks or keypoints for each frame of each video is possible, it is an expensive and time-consuming affair.
We circumvent this through a bootstrapped approach.
For baby segmentation masks, we fine-tune a PointRend segmentation model~\cite{kirillov2020pointrend} on $\sim$500 videos with 10 linearly spaced frames from each.
Similarly, for keypoints, we finetune HRNet~\cite{wang2020HRnet} on $\sim$1500 videos with 20 linearly spaced frames from each.
We apply both models to all video frames of the training set and use the predictions as pseudo-labels during multi-task training.

Our complete multi-task model has a segmentation head, a keypoint estimation head, and all the other anthropometric regression heads (see Fig.~\ref{fig:method}).
We use a Fully Convolutional Network (FCN) head~\cite{long2015fcn} to perform segmentation since we do not need fine precision.
For keypoint estimation, we use a simple 2-layer MLP that regresses the spatial coordinates of keypoints from each frame embedding $\bx_l$.
The model is trained end-to-end through a combination of all losses:
{\small
\begin{equation}
L_\text{total} = L_\text{anthro} + \lambda_m \sum_l L_m(m'_l, \hat{m}_l) + \lambda_k \sum_l L_k(k'_l, \hat{k}_l) \, ,
\end{equation}}
where $m'_l$ and $k'_l$ are the frame-level segmentation mask and keypoints generated by our multi-task model,
$\hat{m}_l$ and $\hat{k}_l$ are pseudo-labels for the mask and keypoints,
$L_m$ is Dice loss~\cite{sudre2017diceloss},
$L_k$ is L1 loss used for keypoints,
and $\lambda_m$ and $\lambda_k$ are loss weights for masks and keypoints.
During inference, we drop both the proxy heads.

\subsection{Augmenting with Tabular Information}
\label{subsec:method:tabular}

The weight of a baby reduces immediately after birth, recovers around days 7-10, and then follows a mostly linear growth trend~\cite{ditomasso2019systematic}.
Hence, knowing the birth weight $w^0$ and current age $a$ is often useful. 
We incorporate this meta information by normalizing them in the $[0, 1]$ range and concatenating them to the output of the pooling layer.
Our final weight regression head (similar to other measurements) is:
\begin{equation}
w = \MLP_w ( [\bz, w^0, a] ) \, .
\end{equation}
We refer to this visual model augmented with tabular features as \textit{W-\model{}} when used to only estimate weight and \textit{\model{}} when used to estimate all anthropometric measures in a multi-task setting.

%% file: tables/anthropometric_correlations.tex
\begin{table}[t]
\small
\centering
\begin{tabular}{c c c c c c}
\toprule
$r(w,l)$ & $r(w,h)$  & $r(w,c)$ & $r(l,h)$ & $r(l,c)$ & $r(h,c)$\\
0.7574 & 0.7176 & 0.7743 & 0.5747 & 0.5901 & 0.7340 \\
\bottomrule
\end{tabular}
\vspace{-2mm}
\caption{We observe high correlation (Pearson correlation coefficient) across anthropometric
measurements on the training set:
weight $w$, length $l$, head circumference $h$, chest circumference $c$.}
\label{table:anthropometric_correlations}
\vspace{-4mm}
\end{table}

%% file: sec/4_experiment.tex
\section{Experiments}
\label{sec:experiments}
We now present empirical validation of our approach on a large dataset collected in community deployment settings.

\input{tables/data_stats}

\subsection{Setup}
\label{subsec:exp:setup}

\paragraph{Dataset collection.}
The data has been collected by 28 trained personnel from rural home settings across 2 geographically diverse regions.
They typically use Android smartphones with a 2-5 MP camera, with cost under \$150.

Our dataset consists of 3439 newborns that are visited on average 3.75 times in the first 42 days of life.
At each visit, we capture three videos with different reference objects (only one is used for one experiment).
While each visit is treated independently in the context of training and evaluation, all visits of a newborn are in the same split.
Ethics committee approvals were obtained prior to data collection and the process itself involves taking informed consent, capturing videos, measuring the weight and anthropometry for the newborn, and providing home-based care recommendations if the newborn is not doing well.
See \cref{sec:app:data_collection} for more details.

\begin{figure*}[t]
\centering
\includegraphics[width=0.33\linewidth]{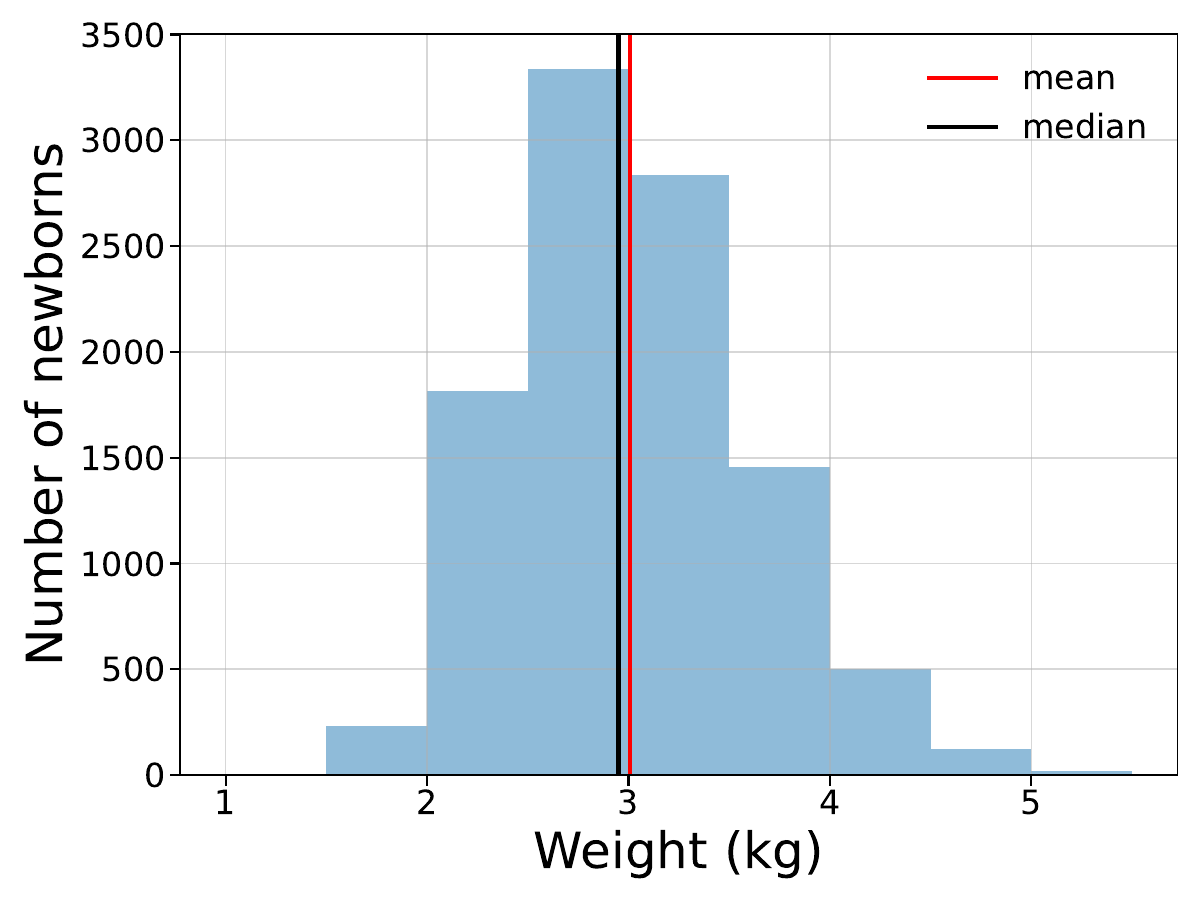}
\hfill
\includegraphics[width=0.33\linewidth]{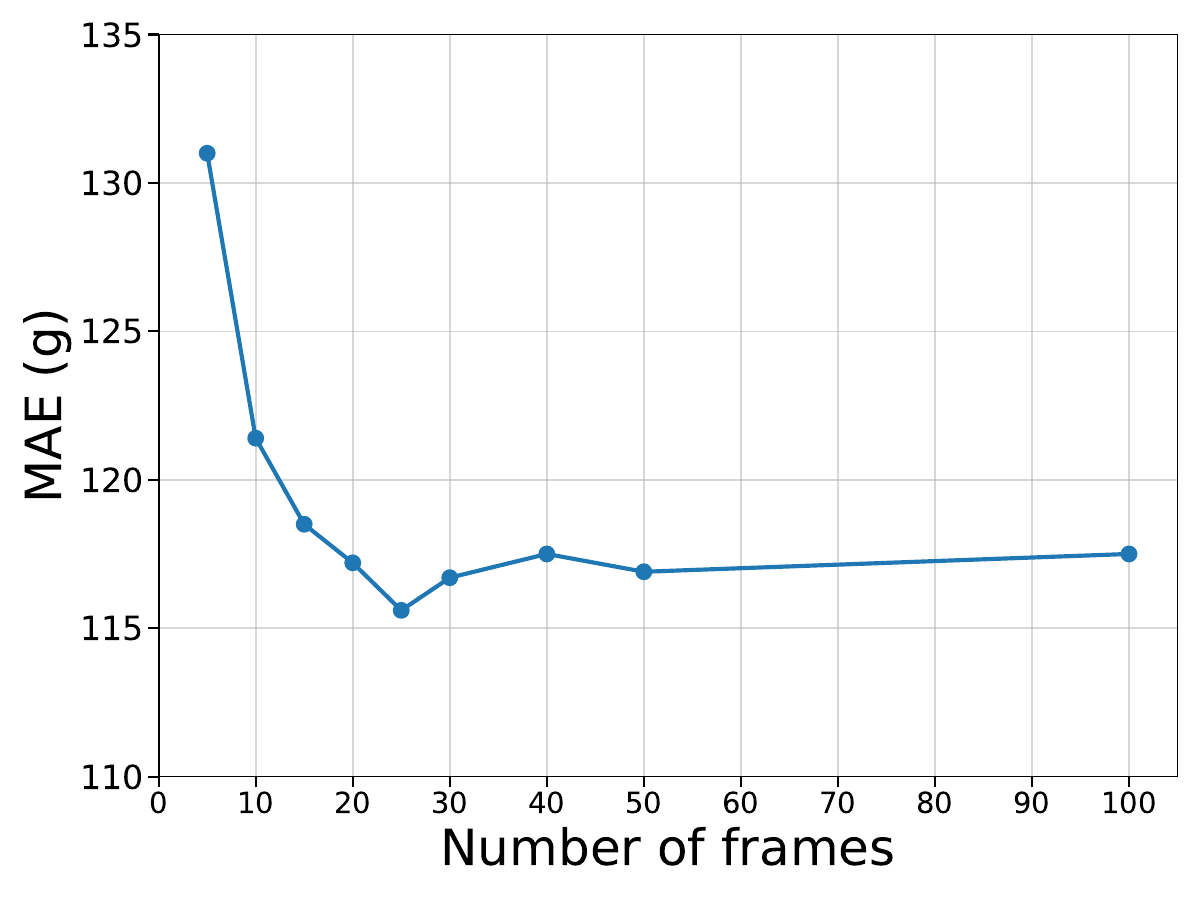}
\hfill
\includegraphics[width=0.327\linewidth]{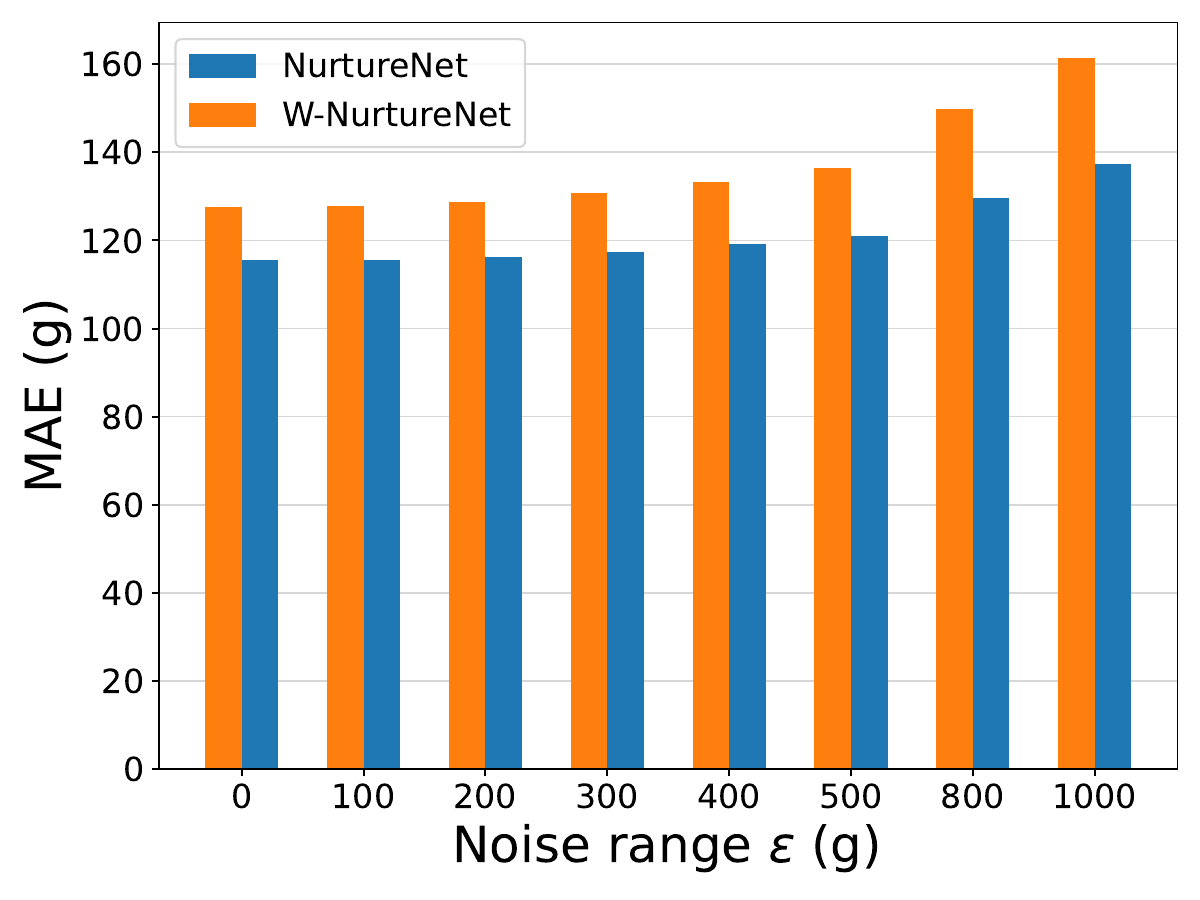}
\vspace{-6mm}
\caption{\textbf{Left:} Weight distribution for the training set.
\textbf{Middle:} Impact of varying the number of frames $N$ during evaluation on the validation set. For training, we use $N{=}25$. The model used here is \model{}, that augments video information with tabular data and uses proxy tasks of baby segmentation mask and keypoints.
\textbf{Right:} Effect on weight MAE on the validation set when adding noise sampled from a uniform distribution to the birth weight for \model{} models.
}
\vspace{-2mm}
\label{fig:multifig}
\end{figure*}

We split the dataset into train (80\%), validation (10\%) and test (10\%), 
while ensuring all visits of a baby are included in the same split.
Table~\ref{table:data_stats} shares the demographics of the dataset.
The weight distribution across splits is matched to the overall dataset distribution (Fig.~\ref{fig:multifig}~(Left)).

\paragraph{Ground-truth.}
We obtain accurate ground-truth weight readings by using a calibrated digital weighing machine (least count \SI{10}{\gram}).
The machine is robust to newborn movement as it stabilizes and locks the recorded value.
A calibrated infantometer is used to measure the length of the newborn, and tape measures are used for measuring the head and chest circumference.

\paragraph{Evaluation metrics.}
We adopt the Mean Absolute Error (MAE) as our primary evaluation metric.
We also report Balanced MAE (BMAE) that averages MAE across each bin of interest as the weight distribution is non-uniform.
Finally, as an error of \SI{200}{\gram} for a \SI{2}{\kilo\gram} newborn is worse than that for a \SI{4}{\kilo\gram} baby, we report relative errors $|w~-~\wgt|/\wgt$ to highlight errors for newborns with lower weights.

\paragraph{Implementation details.}
We fine-tune a ResNet-50 CNN~\cite{he2016resnet} to produce $d {=} 2048$ dimensional representations $\bx_l$ for each frame $f_l$.
We choose $N {=} 25$ frames from a video with an average duration of \SI{12}{\second}.
The selected frames are padded to form a square and resized to 224$\times$224$\times$3.
Augmentations such as vertical / horizontal flips, translations, and color jitter are applied during training.
We use the Adam optimizer~\cite{KingmaB14} and train the model for 200 epochs.
When not mentioned otherwise, the initial learning rate is $10^{-5}$.
We use a StepLR scheduler wherein the learning rate steps down by a factor of 2 every 50 epochs.

\subsection{Model Ablations}
\label{subsec:exp:ablations}

\input{tables/video_based}

\paragraph{Frame- \vs~Video-based weight estimation.}
In the video-based weight estimation approach (Sec.~\ref{subsec:method:video-based_weight}), we fuse $N$ representations early on in the network.
An alternative frame-based approach would be to use individual frames to obtain weight estimates and average over multiple frames during inference (similar to~\cite{shu2023single}).
A constant learning rate of $10^{-5}$ works best for frame-based models.

Table~\ref{table:video_based} shows that video-based models outperform frame-based by a large margin: reduce MAE by \SI{66}{\gram}.
In particular, we also observe that max-pooling outperforms average pooling and vanilla self-attention (SA)~\cite{bahadanau2014attention}.

A key hyperparameter for video-based models is the number of sampled frames $N$. 
Fig.~\ref{fig:multifig}~(Middle) shows that the weight MAE reduces dramatically with increasing $N$, reaches the minimum around $N{=}25$ and slightly increases thereafter.
We choose $N{=}25$ for the rest of the experiments.

\paragraph{Impact of ResNet-50 parameters.}
We experiment with a ResNet-50 encoder pretrained on the ImageNet (IN) dataset and one pretrained using the Contrastive Language-Image Pretraining (CLIP) technique~\cite{radford2021learning}.
For IN models, an initial learning rate of $10^{-4}$ stepped down by a factor of 2 every 30 epochs works best.
As seen in Table~\ref{table:video_based}, CLIP-based CNN initialization results in better performance, and this encoder is used in all further experiments.

\paragraph{Impact of frame sampling.}
The frame selection process influences the representation and the weight estimate.
To reduce this dependency, we introduce subsampling as an augmentation during training.
Specifically, we randomly pick $N'{=}40$ frames, and subsample 10 subsets of $N{=}25$ frames with replacement.
By requiring the model to produce the same estimate (Eq.~\ref{eq:mlpw}) across these subsets, we make the model less sensitive to frame selection.
During inference, we do not use subsampling.
Table~\ref{table:video_based} shows that subsampling typically results in a modest improvement of \SI{5.5}{\gram}.
We employ this technique for further experiments.

\paragraph{Impact of reference object.}
Table~\ref{table:video_based} shows that using \faChessBoard{}, a standard object for camera calibration, as a reference object over the \faRuler{} gives a \SI{10.9}{\gram} improvement on MAE.
However, for wider adoption and given \faRuler's availability in resource constrained areas, we restrict our solution to using a \faRuler{}.

\input{tables/video_multitask}
\paragraph{Impact of multi-task approaches.}
Table~\ref{table:video_multitask} shows the results of combining various tasks, and the corresponding MAE.
As a simple baseline, row 0 displays performance for using the mean value of the training set as the prediction.
In all experiments, the loss coefficients are set to
$\lambda_w {=} 5.0,
\lambda_l {=} 0.1,
\lambda_h {=} 0.1,
\lambda_c {=} 0.1$,
$\lambda_m {=} 3.0$, and
$\lambda_k {=} 100.0$,
to scale the importance of various losses.
Rows 1-4 show the results when performing each anthropometry estimation task independently, indicating that they fare much better to not using any model.
Row 5 shows the effect of including proxy visual tasks, which leads to a performance gain of $\SI{16.5}{\gram}$.
Row 6, compared to rows 1-4, shows negligible change.
This indicates that we can use one multi-task model that estimates all anthropometric measures with one video.
Finally, row 7, combines all tasks showing marked improvement across all measurements.

\input{tables/vision_tabular}
\paragraph{Tabular only model}
Tab-Only is posed as a simple linear regression that takes in 2 inputs (birth weight $w^0$ and age~$a$) and predicts the current weight $w$.
Row 1 in Table~\ref{table:vision_tabular} shows that this model achieves a competitive MAE of \SI{202.5}{\gram}.
However, this model is not useful in practice as it predicts the same weight for all babies with a given birth weight after $a$ days,
\ie~this model cannot predict deviations from the mean growth for the population (training set).

\paragraph{Augmenting visual information with tabular inputs.}
We augment our visual representation $\bz$ by concatenating them with $[0, 1]$ normalized tabular inputs and send them forward to regress weight
(W-\model{}, Sec.~\ref{subsec:method:tabular}).
Row 2 of Table~\ref{table:vision_tabular} shows that W-\model{} achieves an impressive \SI{12.2}{\gram} improvement in MAE from \SI{139.9}{\gram} to \SI{127.7}{\gram}.
Rows 3 and 4 show results on training the multi-task \model{}.
While row 3 shows a \SI{14}{\gram} improvement in MAE on using auxiliary tasks, row 4 is a unified model that can estimate all anthropometry measurements and shows a \SI{12}{\gram} improvement in weight MAE.

\paragraph{Effect of errors in recorded birth weight.}
To deploy models in rural settings, an important factor to consider is the erroneous nature of tabular inputs.
We simulate these errors as $\tilde{w}^0 = w^0 + \epsilon$,
where the noise $\epsilon$ is sampled from a uniform distribution $\mcU(-q, q)$ and $q$ corresponds to the maximum deviation in \unit{\kilo\gram}.
Fig.~\ref{fig:multifig}~(Right) shows that our models are quite robust to noisy inputs.
In fact, when $q {=} $\SI{0.5}{\kilo\gram},
W-\model{} achieves an MAE of \SI{136.5}{\gram} (worse than when $q{=}0$ by \SI{9}{\gram}), still better than our video-based model at \SI{139.9}{\gram}.
\model{} is more robust to noise than W-\model{} and results in a $\sim$\SI{5}{\gram} increase in MAE.
Finally, the Tab-Only model is highly sensitive to errors in birth weight and results in an MAE of \SI{307.4}{\gram} (up by \SI{105}{\gram}).

\input{tables/baseline_ablations}

\subsection{Baselines, Results Summary}
\label{subsec:exp:baselines}
We now present and evaluate a few baselines for weight estimation:
(i)~A naïve approach is to predict the mean of the training set. This acts like the upper bound of the error for any model.
(ii)~A second approach uses structural information that can be extracted from predicted segmentation masks of the newborn and the ruler.
We extract hand-crafted representations in the form of Hu image moments~\cite{hu1962visual} or region features~\cite{burger2009principles}.
(iii)~We also evaluate the Histogram of Oriented Gradients (HOG) features~\cite{dalal2005histograms} that are popular in classical computer vision literature.
RBF-kernel Support Vector Regressors (SVR)~\cite{smola2004tutorial} are used to obtain anthropometry estimates from all three representations.

\paragraph{Baseline ablations.}
Table~\ref{table:baseline_ablations} shows the MAE for weight estimation for all three feature representations and a combination.
Regionprops features, together with Hu moments show best performance for weight estimation (\SI{393.0}{\gram}).
Here as well, the reference object is useful and computing features from both the baby and ruler regions improves performance.
\cref{sec:app:baselines}
presents additional details on features and ablations.
However, performance of all baselines is far from proposed deep-learning based approaches.

\input{tables/best_results_on_test}
\paragraph{Comparing best approaches.}
We summarize the key methods on the test set in Table~\ref{table:best_results_on_test}.
In particular, we observe that video-based models (\SI{139.0}{\gram}) achieve a large improvement over the best hand-crafted representations (\SI{390.1}{\gram}) and single frame based approach~\cite{shu2023single} (\SI{214.0}{\gram}).
Augmenting the video-only models with tabular data improves MAE to \SI{126.4}{\gram}. 
Finally, \model{} results in best performance, achieving an MAE of \SI{114.3}{\gram}, while presenting a unified model for newborn anthropometry.
Table~\ref{table:compression} (Uncompressed) presents results on all measures of \model{}.

\subsection{Analysis and Discussion}
\label{subsec:exp:analysis}

\input{tables/slice_dice_weight}

\paragraph{Results sliced by weight}
bins are presented in Table~\ref{table:slice_dice_weight}.
\model{} brings large improvements in MAE from \mySIrange{15}{80}{\gram} across all bins, but particularly for the low \mySIrange{1}{2}{\kilo\gram} and high \mySIrange{4}{5}{\kilo\gram} bins.
Inclusion of the multi-task approach and the tabular inputs also reduces relative and $80^\text{th}$ percentile errors across all bins.
Low errors in the \mySIrange{1}{2.5}{\kilo\gram} bins are especially important to identify underweight newborns.

\begin{figure}[t]
\begin{minipage}[c]{0.66\linewidth}
\includegraphics[width=0.99\linewidth]{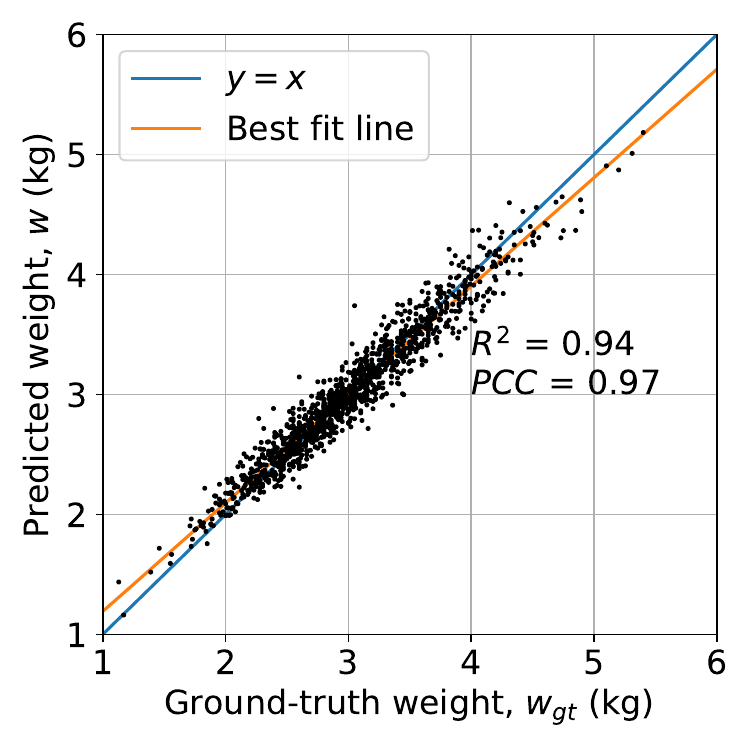}
\end{minipage}
\begin{minipage}[c]{0.32\linewidth}
\caption{
Scatter plot showing predicted weight \vs~ground-truth weight for \model{} on the test set.
The best fit line (least squares) lies close to the $y {=} x$ diagonal, indicating the goodness of our model.
$R^2$ is the coefficient of determination and $PCC$ is the Pearson correlation coefficient.
} \label{fig:pred-gt}
\end{minipage}
\end{figure}

\paragraph{Predictions \vs Ground-truth}
Another way to assess the performance of our proposed model is a scatter plot of the model’s predictions against the ground-truth.
Fig.~\ref{fig:pred-gt} shows the scatter plot for \model{} on the test set.
While we observe that the best fit line is close to the $y{=}x$ diagonal, our model tends to slightly over-predict for low weight and under-predict for higher weight newborns.
This can be attributed to the dataset imbalance (Fig.~\ref{fig:multifig}~(Left)).

\cref{sec:app:experiments} presents tSNE plots by weight, age, and data collector; and Bland-Altman plot for weight measurements.

\input{tables/compression}

\paragraph{Model compression.}
Training our model requires 
one V100 GPU with \SI{32}{\giga\byte} memory.
However, our goal is to deploy the \model{} model 
on a low-cost smartphone that can be used by health workers in underserved geographies.
Furthermore, the lack of internet coverage necessitates a drastic reduction in the memory and computational footprint of the model to enable on-device and offline inference.
We prune \model{} using the NNI library~\cite{nni2021}.
Specifically, we use the L1NormPruner twice to discard half the output channels having the smallest L1 norm of weights in each iteration, effectively reducing the size of the compute requirement by 75\%.
Further, we perform static quantization, converting the FP32 weights and activations to INT8.
The result is a model that is 8$\times$ smaller and 4$\times$ faster with an acceptable deterioration in performance up to \SI{3.5}{\gram} MAE.
Table~\ref{table:compression} shows that compression leads to a negligible performance drop for weight (W), head circumference (H), and chest circumference (C), but an acceptable increase in length (L) error.

\paragraph{\model{} \vs Conventional practice.}
We conduct a preliminary and independent field study to analyze the errors in weight measurements made through conventional practices and compare them against \model{}.
Weight readings taken by health workers using spring balances are compared against calibrated digital weighing machines used for ground-truth weight measurements.
We observe an MAE of \SI{183}{\gram} (N=92) for conventional methods indicating the challenges of recording such data in rural community settings.
Note, that this result is biased towards being lower as the health workers knew that they were being monitored and can only be expected to be worse in real scenarios.
\model{} achieves a lower MAE at \SI{114.3}{\gram} (N=1295), indicating the field-readiness of our approach.

\paragraph{Limitations.}
While AI models can provide meaningful accuracy in many cases, they cannot be perfect on all samples, particularly when it comes to complex problems like estimating the weight of a baby.
This may be due to various factors such as newborn clothing, lighting conditions, environmental conditions, camera angles, the position of the baby relative to the camera, or the baby's movements.
The model may also struggle to accurately estimate the weight of babies with certain physical characteristics (\eg~missing limbs) or rare medical conditions that affect growth.

%% file: tables/data_stats.tex
\begin{table}[t]
\small
\captionsetup{skip=2mm, font=small}
\centering
\begin{tabular}{l ccc ccc}
\toprule
& \multicolumn{3}{c}{Visits} & \multicolumn{3}{c}{Newborns} \\
Source & Train & Val & Test & Train & Val & Test \\
\midrule
Region 1 & 8735 & 1096 & 1075 &  2304 & 293 & 280 \\
Region 2 & 1590 & 185 & 220 &  447 &  51 &  64 \\
\midrule
Total  & 10325 & 1281 & 1295 & 2751 & 344 & 344 \\
\bottomrule
\end{tabular}
\vspace{-1mm}
\caption{Number of visits and newborns from rural home settings.}
\vspace{-4mm}
\label{table:data_stats}
\end{table}

%% file: tables/video_based.tex
\begin{table}[t]
\small
\captionsetup{skip=2mm, font=small}
\centering
\tabcolsep=0.08cm
\begin{tabular}{cc ccc cc}
\toprule
Input & Backbone & RefObj & SS & Pooling & MAE (\SI{}{\gram}) & BMAE (\SI{}{\gram})\\
\midrule
\multirow{1}{*}{Frame} & IN & \faRuler & $-$ & $-$ & 224.6 & 437.2\\
\midrule
\multirow{3}{*}{Video}
& IN &\faRuler & $-$ & Average & 170.0 & 290.6\\
& IN &\faRuler & $-$ & Vanilla SA & 167.1 & 291.9\\
& IN &\faRuler & $-$ & Max & 158.5 & 223.7\\
\midrule
\multirow{1}{*}{Video}
& CLIP &\faRuler & $-$ & Max & 145.4 & 207.2\\
\midrule
\multirow{2}{*}{Video}
& CLIP &\faRuler & \checkmark & Max & 139.9 & 211.3\\
& CLIP &\faChessBoard & \checkmark & Max & 129.0 & 189.0\\
\bottomrule
\end{tabular}
\caption{Impact of input, reference object (chessboard \faChessBoard, and wooden ruler \faRuler), frame subsampling (SS), and pooling methods.
MAE and BMAE are reported for weight estimation on the validation set.
The backbone is ResNet-50 pretrained either on ImageNet (IN)~\cite{he2016resnet} or CLIP~\cite{radford2021learning}.
We see consistent performance improvement across all modifications -- video-based models, CLIP as backbone, and including sub-sampling.
} 
\vspace{-4mm}
\label{table:video_based}
\end{table}

%% file: tables/video_multitask.tex
\begin{table}[t]
\small
\captionsetup{skip=2mm, font=small}
\centering
\tabcolsep=0.12cm
\begin{tabular}{c cccccc cccc}
\toprule
& \multicolumn{6}{c}{Tasks} & \multicolumn{4}{c}{MAE} \\
\cmidrule(lr){2-7}\cmidrule(lr){8-11}
& W & L & H & C & M & K &
W (\unit{\gram}) & L (\unit{\centi\meter}) & H (\unit{\centi\meter}) & C (\unit{\centi\meter}) \\
\midrule
0 & \multicolumn{6}{c}{Train set mean} & 495.8 & 2.36 & 1.80 & 2.38 \\
\midrule
1 & \checkmark & - & -  & - & - & - & 139.9 &  $-$   & $-$ & $-$ \\
2 & - & \checkmark & -  & - & - & - & $-$ &  1.53   & $-$ & $-$ \\
3 & - & - & \checkmark  & - & - & - & $-$ &  $-$   & 1.15 & $-$ \\
4 & - & - & -  & \checkmark & - & - & $-$ &  $-$   & $-$ & 1.37 \\
\midrule
5 & \checkmark & - & -  & - & \checkmark & \checkmark & 123.4 &  $-$   & $-$ & $-$\\
6 & \checkmark & \checkmark & \checkmark & \checkmark & - & - & 138.2 &  1.51   & 1.13 & 1.37\\
7 & \checkmark & \checkmark & \checkmark  & \checkmark & \checkmark & \checkmark & 124.4 & 1.39 & 1.08 & 1.27\\
\bottomrule 
\end{tabular}
\caption{
Video-based model with CLIP backbone, \faRuler{}, Max pooling, and frame subsampling, under different multi-task configurations.
W: weight,
L: length,
H: head circumference,
C: chest circumference,
M: segmentation mask,
K: keypoints.
Performance on the validation set improves as we incorporate all tasks.
}
\vspace{-1mm}
\label{table:video_multitask}
\end{table}

%% file: tables/vision_tabular.tex
\begin{table}[t]
\small
\captionsetup{skip=2mm, font=small}
\centering
\tabcolsep=0.10cm
\begin{tabular}{l l c c c}
\toprule
& \multirow{2}{*}{Model} & \multirow{2}{*}{Tasks} & \multicolumn{2}{c}{Weight estimation} \\
& & & MAE (\SI{}{\gram}) & BMAE (\SI{}{\gram}) \\
\midrule
0 & Video-only & W & 139.9 & 211.3 \\ 
1 & Tab-only &  W  & 202.5 & 322.6 \\
2 & W-\model{} & W & 127.7 & 196.6 \\
\midrule
3 & \model{} & W M K & 113.7 & 171.8 \\
4 & \model{} & W L H C M K & 115.6 & 181.3 \\
\bottomrule 
\end{tabular}
\caption{
W-\model{} concatenates the visual representation to the tabular inputs to regress weight to show improved performance (validation set). \model{} is the multi-task equivalent. The Tasks column shows the set of tasks on which the model is trained.
}
\vspace{-3mm}
\label{table:vision_tabular}
\end{table}

%% file: tables/baseline_ablations.tex
\begin{table}[t]
\small
\centering
\tabcolsep=0.08cm
\begin{tabular}{l c c c c c}
\toprule
& Train & Hu + & HOG & Regionprops & Hu + Region-\\
& set mean & SVR & + SVR & + SVR & props + SVR \\
\midrule
W (g) & 495.8 & 470.3 & 466.7 & 399.4 & 393.0 \\
\bottomrule
\end{tabular}
\vspace{-2mm}
\caption{Hand-crafted models perform poorly on weight regression (validation set) compared to proposed models.}
\label{table:baseline_ablations}
\vspace{-3mm}
\end{table}

%% file: tables/best_results_on_test.tex
\begin{table}[t]
\small
\centering
\begin{tabular}{l c c}
\toprule
Method& MAE (\SI{}{\gram})& BMAE (\SI{}{\gram})\\
\midrule
Train set mean & 483.5 & 1091.8 \\
Best hand-crafted approach & 390.1 & 716.7 \\
Frame-based method & 214.0 & 409.7 \\
Best video-only model & 139.0 & 222.5 \\
W-\model{} & 126.4 & 207.1\\
\model{} (W L H C M K) & \bf{114.3} & \bf{157.5} \\
\bottomrule
\end{tabular}
\vspace{-2mm}
\caption{Weight estimation performance on the test set.}
\vspace{-2mm}
\label{table:best_results_on_test}
\end{table}

%% file: tables/slice_dice_weight.tex
\begin{table}[t]
\footnotesize
\centering
\captionsetup{skip=2mm, font=small}
\tabcolsep=0.08cm
\begin{tabular}{c c ccc ccc}
\toprule
Weight & Test Set &
\multicolumn{3}{c}{Video-based model} & 
\multicolumn{3}{c}{\model{}} \\
Bin (\SI{}{\kilo\gram}) & Count & MAE (\SI{}{\gram}) & E80 (\SI{}{\gram}) & \% Rel & MAE (\SI{}{\gram}) & E80 (\SI{}{\gram}) & \% Rel \\
\midrule
1 - 2   & 40  & 207.5 & 305.1 & 12.2 & 127.3 & 180.9 & 7.3 \\
2 - 2.5 & 226 & 135.2 & 218.4 &  5.9 & 108.1 & 170.5 & 4.7 \\
2.5 - 3 & 420 & 111.5 & 177.7 &  4.1 & 97.7 & 160.0 & 3.6 \\
3 - 3.5 & 353 & 137.2 & 224.4 &  4.2 & 115.1 & 181.8 & 3.6 \\
3.5 - 4 & 179  & 148.0 & 229.3 &  4.0 & 123.5 & 197.6 & 3.3 \\
4 - 5   & 73  & 245.4 & 425.8 &  5.7 & 187.6 & 288.5 & 4.4 \\
\midrule
All     & 1291 & 139.0 & 223.2 &  4.8 & 114.3 & 179.8 & 3.9 \\
\bottomrule
\end{tabular}
\caption{Results sliced by weight bins on the test set.
Metrics: E80 corresponds to the $80^\text{th}$ percentile absolute error and indicates that 80\% of the samples have an error less than this value.
\%~Rel corresponds to mean absolute relative error.
\model{} improves over the video-based model across all weight bins.}
\label{table:slice_dice_weight}
\vspace{-4mm}
\end{table}

%% file: tables/compression.tex
\begin{table}[t]
\small
\tabcolsep=0.12cm
\centering
\begin{tabular}{ccccccc}
\toprule
\multirow{2}{*}{Model} & \multirow{2}{*}{Size (MB)} & \multirow{2}{*}{GFLOPs} & \multicolumn{4}{c}{MAE (\unit{\gram}, \unit{\centi\meter}, \unit{\centi\meter}, \si{\centi\meter})} \\
\cmidrule(lr){4-7}
& & & W & L & H & C \\
\midrule
Uncompressed & 121.6 & 5.38 & 114.3 & 1.33 & 1.04 & 1.25\\
Pruned & 30.4 & 1.35 & 116.0 & 1.26 & 1.02 & 1.21\\
Quantized & 15.0 & 1.35 & 117.8 & 1.53 & 1.05 & 1.26 \\
\bottomrule
\end{tabular}
\vspace{-2mm}
\caption{Performance of the uncompressed \model{}, pruned, and quantized models on the test set.
We are able to reduce models by 8$\times$ with minimal loss in weight estimation performance.
}
\vspace{-4mm}
\label{table:compression}
\end{table}

%% file: sec/5_conclusion.tex
\section{Conclusion}
We present a vision system for newborn anthropometry from a short video taken with a low-cost smartphone.
Our proposed approach computes a video representation and augments it with tabular data to obtain weight estimates.
We extend this model through multi-task training to simultaneously estimate other anthropometric measurements such as the length, head circumference, and chest circumference (\model). 
A proxy task of predicting baby segmentation masks and keypoints further improves the weight estimation performance.
Using pruning and quantization, we compress \model{} to \SI{15}{\mega\byte}, allowing offline inference and deployment on low-cost smartphones.

This solution is envisioned as a public health screening tool and is currently not intended for diagnostic or clinical settings where good anthropometric instruments and trained personnel are available.
Such a tool provides a convenient, geo-tagged, and contactless way for health workers and public health systems to monitor the growth and development of newborns, enabling targeted interventions to drive better health outcomes.

{\small
\paragraph{Acknowledgments.}
We thank the Bill \& Melinda Gates Foundation (BMGF) and the Fondation Botnar for supporting this work.
We thank Niloufer Hospital at Hyderabad for initial data collection.
We are grateful to SEWA Rural at Gujarat and PGIMER at Chandigarh for collecting data in community settings.
We thank all members of WIAI who were involved for their contributions.
}

%% file: sec/6_suppl.tex
\section*{Appendix}

We present additional details with regard to 
the data collection procedure (Sec.~\ref{sec:app:data_collection}) and
the data validation process to ensure that the models see correct inputs (Sec.~\ref{sec:app:data_validation}).
Next, we present details related to the experiments:
Sec.~\ref{sec:app:baselines} discusses the baseline approach, while 
Sec.~\ref{sec:app:experiments} presents clarification regarding metrics and further analysis.

\appendix
\section{Data Collection Process}
\label{sec:app:data_collection}
As described in the main paper, each baby is visited multiple times in the first 6 weeks of life.
The data collector visits and captures videos of the baby around the 3, 7, 14, 21, 28, and 42 days after birth to match the health program's recommended schedule.
However, due to field and logistical challenges, we do encourage the data collector to visit the newborn within a $\pm$2 day window.
This gives us an average of 3.75 visits per newborn.

\begin{figure}[t]
\centering
\includegraphics[width=0.65\linewidth]{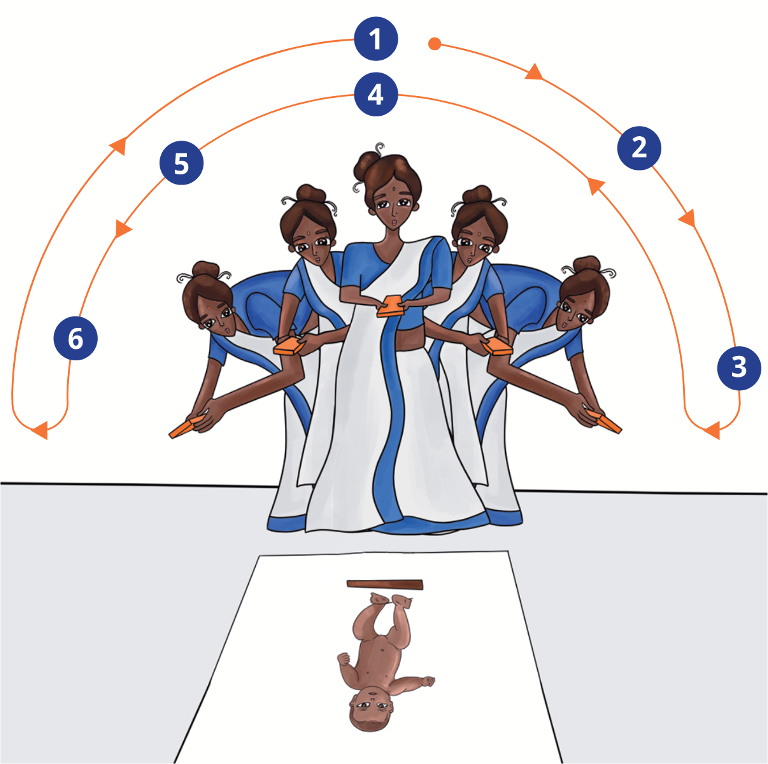}
\caption{Video recording process followed by health workers to capture the newborn from multiple viewing angles.}
\vspace{-4mm}
\label{fig:video_collection}
\end{figure}

The data collectors are trained to capture a video by starting from the top of the baby and making a smooth arc as illustrated in \cref{fig:video_collection}.

\paragraph{Enrolment.}
At the first visit, the baby is enrolled using a custom-developed mobile application to ensure data security.
The application generates automatic reminders for the data collector to do follow-up visits.
Prior to enrolment, the data collectors explain the project to the parents and obtain their informed consent in the local language.
During enrolment, we capture basic information such as the mother's and newborn's name, address, sex, mode of delivery, date of birth, and weight at birth.

\paragraph{At each visit}
the data collectors are trained to adhere to the following protocol:
\begin{enumerate}[nosep]
\item After greeting the parents, the first task is to setup the video capture environment: find a flat, well-lit area in the house, arrange for a bedsheet on which the baby will be placed, and prepare the reference objects.
\item Next, the digital weighing machine is prepared for measuring ground-truth.
The baby is brought in and it's clothes are removed. The newborn is successively placed three times on the weighing machine and readings are noted for each measurement.
The whole process is captured in a video to ensure adherence to protocol (see Sec.~\ref{subsec:app:gtweight}).
As indicated in the main paper, we ensure high quality ground-truth (\SI{10}{\gram} least count) by using a custom-built, calibrated, and certified
weighing machine that averages weight over time.
\item We then capture three videos of the baby with different reference object conditions: no reference object, chessboard (\faChessBoard), and the wooden ruler (\faRuler).
For each video, the data collector places the appropriate reference object and makes an arc around the baby as indicated in Fig.~2 of the main paper.
We attempt to capture the newborn's shape
by making a steady arc around it while ensuring minimal motion blur (due to camera motion)
and that the newborn and the reference object are in the field of view at all times.
\item The data collector also measures the newborn's length using an infantometer, and its head and chest circumference using tape measures.
We train our models in a multi-task manner to predict these measurements.
\item Finally, an oral health assessment is performed by quizzing the parents on aspects such as feeding status, breathing rate, appearance, muscle tone, and discharge from the eyes or umbilicus.
\item In case of any concerns or anomalous responses, the data collectors counsel the parents on potential recourses to address them.
\end{enumerate}

The data is automatically synced to secure cloud storage when the mobile device has access to the internet (note that rural areas where data collection happens may not necessarily have access to the internet) and de-identified before sharing for further processing.

\section{Data Validation Criteria}
\label{sec:app:data_validation}

We are interested in understanding the data quality through various annotations related to the environment, the use of appropriate reference object, clothing artifacts on the newborn, and ground-truth. 
We obtained videos of 16,612 visits across two geographically diverse regions.
A team of 5 annotators was trained on the prescribed protocol, and 2 annotators independently annotated each video.
After validation, we were left with 12,901 usable visits.
Table~\ref{table:data_validation} enlists the criteria used to discard visits.
This validation protocol involves three sequential steps as described in the subsections below.

\subsection{Environment Validation}
\label{subsec:app:video_object_validation}

Our data is collected in everyday houses in rural, low resource areas in low- and middle-income countries (LMICs) where the video capture environment is unconstrained.
This leads to diverse variations in the visual settings across the captured videos and props up classic vision challenges related to poor lighting; bedsheets of different colors, shapes, and textures; and other challenges related to data collection, such as the lack of a video capture setup potentially leading to motion blur and inconsistency in recorded videos.
This is far from clinical settings (\eg~a hospital) where all newborns may be brought to the same room or even the same bed and captured by the same data collector (nurse) with the same device, making the vision problem easier to solve.

Our first check ensures that each video has a newborn with the correct reference object.
Note that for each visit we collect three videos with different reference object conditions:
no reference object, with a \faChessBoard, and with a \faRuler.
Due to the simple nature of this task, we use unanimity to ensure that the annotations are correct. We remove 53 visits after this check leaving us with 16,559 visits that are passed on to the next stage.

\subsection{Video Quality Validation}
\label{subsec:app:input_validation}
\input{tables/data_validation}
We validate the quality of the videos with the aid of a questionnaire to determine the quality of data collection and ensure adherence to protocol.
The questions are:
(i)~is the newborn wearing clothes?
(ii)~is the newborn cropped?
(iii)~is the reference object cropped?
(iv)~is there good and sufficient light?
(v)~is the video blurry?
(vi)~is the newborn and reference object on the same plane?
(vii)~are there other humans visible in the video?
(viii)~is the arc smooth or jerky? and
(ix)~is the newborn captured well from both left and right side angles (\ie~how complete is the arc)?

We accept partial failures (\eg~newborn cropped for \mySIrange{1}{3}{\second}) in most of the above criteria and observe that complete failures (correspondingly, newborn cropped for $\ge$\SI{3}{\second}) are quite rare.
We plan to use the annotations for future analysis and potential studies in error attribution.
As we are interested in building a robust anthropometry estimation system, we realize that all videos will not be captured well during deployment. 
We discard 441 visits in this process and are left with 16,118 visits.

\subsection{Ground-truth Weight Validation}
\label{subsec:app:gtweight}

\input{tables/weight_validation}

The third and final validation check concerns the ground-truth weight. It involves annotators watching the video recording in which the ground-truth weight of the newborn is captured and recording the observed weight.
Recall that the newborn is placed thrice on the weighing machine leading to a total of 6 weight readings across two annotators.

Visits that have 4 of 6 weight readings in agreement are directly accepted.
Alternatively, if no reading is more than \SI{50}{\gram} away from the mean of all 6 readings, we accept the visit.
All other visits are passed through the criteria below.
We discard the visits if any of the following is true:
(i)~the newborn is not visible on the weighing machine;
(ii)~newborn is wearing clothes while being placed on the machine;
(iii)~readings are not stable with two or more than two readings varying beyond \SI{50}{\gram};
and
(iv)~a large chunk is attributed to other problems such as the weighing machine that may not be placed on a proper flat surface or is not visible in the video due to occlusions or lack of focus or glare, someone's hand touching the weighing pan, the newborn's limbs are touching a nearby wall, \etc.
Visits that do not fail any of the 4 rejection criteria are also accepted.
Table~\ref{table:weight_validation} shows the counts of the rejection criteria where we discard the visits.

The stringent ground-truth weight annotation protocol along with our weighing machine with a hold function
allows us to capture highly accurate values of the ground-truth weight. 
We removed 3200 visits in this process and are left with 12,918 visits. 
17 more visits are finally removed since they have short videos (less than 40 frames as required for subsampling).
Finally, 12,901 videos are used as part of our experiments.

\begin{figure*}[t]
\centering
\captionsetup{skip=0mm, font=small}
\includegraphics[width=0.355\linewidth]{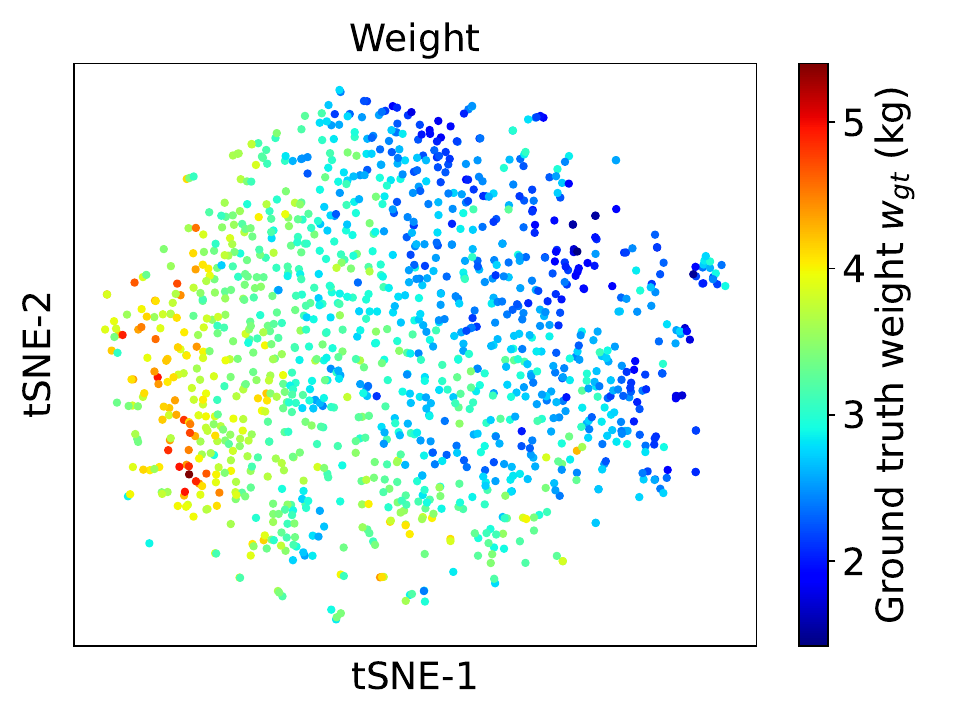}
\includegraphics[width=0.355\linewidth]{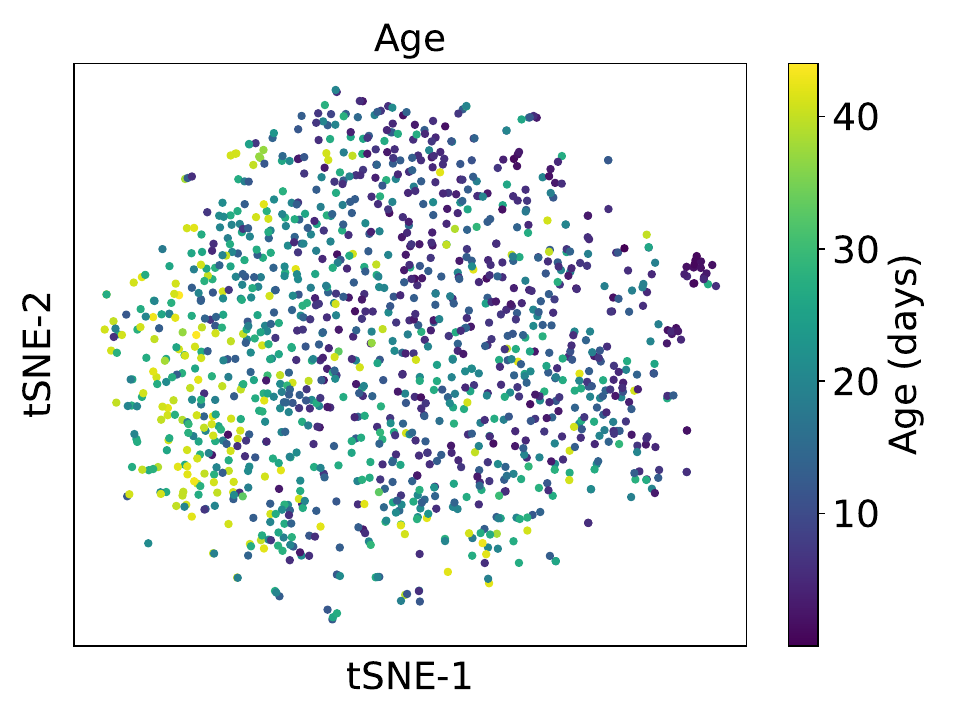}
\includegraphics[width=0.28\linewidth]{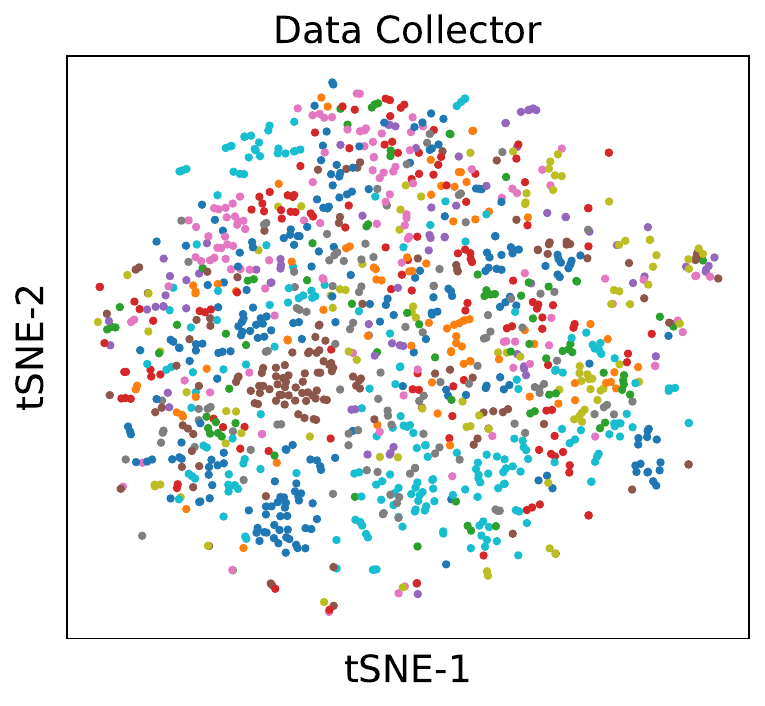}
\vspace{1mm}
\caption{
t-SNE embeddings of representations from the video-based model on the validation set.
Each dot is colored by different properties:
weight (left), age (center), and data collector (right).
}
\label{fig:tsne}
\end{figure*}

\section{Baselines}
\label{sec:app:baselines}
\input{tables/handcrafted_feature_model_info}
\input{tables/baseline_ablations_extended}

For all our baseline experiments, we use \texttt{scikit-image}~\cite{van2014scikit} for extracting the hand-crafted features.
Hu moments and Regionprops are extracted from the binary masks of the baby and wooden ruler regions, while the HOG features are extracted from the combined baby and wooden ruler regions cropped from the original image.
The hand-crafted features across $25$ frames for a given video are concatenated together to create the final feature vector (Table~\ref{table:hand_craft_fea_model_info}).

We evaluate three regression models:
ordinary least squares based Linear Regression (LR),
Multi-Layer Perceptron (MLP), and kernel Support Vector Regressor (SVR). For LR and MLP, we scale the Hu moments with a log transformation to reduce the variability in feature values. The z-score and minmax feature scalers have been experimented with.
For the MLP, we use the \texttt{scikit-learn}~\cite{pedregosa2011scikit} implementation, and for the kernel SVR, we use the \texttt{LIBSVM}~\cite{chang2011libsvm} implementation.
For MLP, we use one hidden layer of 100 units with an initial learning rate of 0.001 and an inverse scaling learning rate scheduler.
For SVR that uses the Radial Basis Function (RBF) kernel, the kernel coefficient $\gamma$ is set to $\frac{1}{d\cdot{\Sigma(X)}}$, where $d$ is the feature dimensionality and $\Sigma(X)$ is the variance of $X$.
Table~\ref{table:baseline_ablations_extended} 
shows the performance of hand-crafted features with different models that regress weight. 
SVR outperforms LR and MLP across all representations with the combined Hu and Regionprops features giving the best performance on weight estimation.

\section{Experimental Details and Analysis}
\label{sec:app:experiments}

We present additional experimental details related to metrics and some analysis.

\subsection{Metric: Balanced MAE}
The standard metric Mean Absolute Error (MAE) 
does not take into account the label distribution (\eg~majority of the newborns in our dataset have weight between \mySIrange{2.5}{3.5}{\kilo\gram}).
As our goal is related to identifying malnutrition in newborns, it is important to get accurate predictions and metrics corresponding to low birth weight newborns.
Thus, we use a more equitable and fair metric: Balanced MAE (BMAE), defined as the average of MAE across multiple weight bins.
In our experiments, we use bins of \SI{500}{\gram} granularity.
Based on the weight distribution in the dataset (see Fig.~\ref{fig:multifig} (Left) in the main paper), we set the lower limit to \SI{1}{\kilo\gram} and the upper limit to \SI{5.5}{\kilo\gram}.
The bins are thus defined as follows:
\begin{equation}
B = \left\{ \left[l, l+0.5\right) \,\, \forall \,\, l \in \{1, 1.5, \ldots, 5\} \right\} \, ,
\end{equation}
and the BMAE metric is defined as:
\begin{equation}
\text{BMAE} = \frac{1}{|B|} \sum_{b \in B} \frac{1}{|b|} \sum_{\wgt \in b} |w - \wgt| \, ,
\end{equation}
where $|B|$ is the number of bins and $|b|$ is the number of samples in a particular bin.

\subsection{t-SNE plots}
Fig.~\ref{fig:tsne} shows t-SNE embeddings~\cite{maarten2008tsne} for videos from the validation set. We use the simple video-based model for this analysis to visualize the feature space to capture the variation of weight without the influence of multiple tasks or tabular information.
(i)~In the left plot, colors indicate the true weight of the newborn in \unit{\kilo\gram}.
We see a smooth color distribution across the embeddings indicating that the model has optimized to a good representation space.
(ii)~The center plot shows the age of the newborn at the time of data collection in days. We observe a smooth transition here as well. However, there are some higher age babies at the top right and vice versa.
(iii)~In the right plot, we color the dots by the data collector. A good mix is observed which is desirable to ensure invariance across data collectors.

\begin{figure}[t]
\centering
\includegraphics[width=1.0\linewidth]{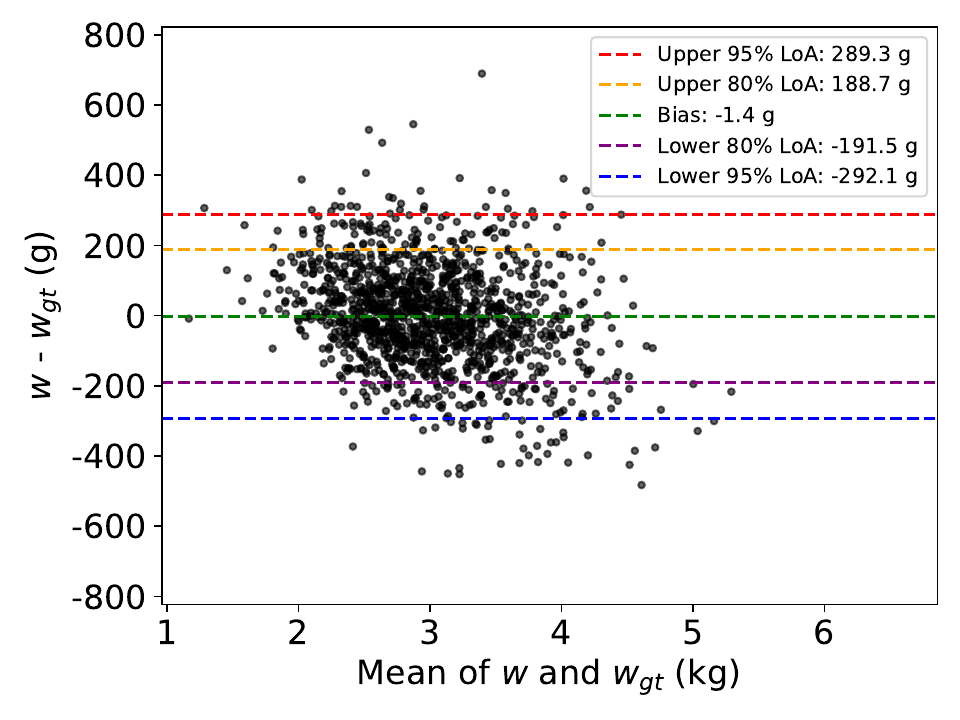}
\vspace{-5mm}
\caption{
Bland-Altman analysis between ground-truth and \model's weight estimates on the test set.
Limits of Agreement (LoA) are plotted at 80\% and 95\% confidence intervals.
}
\vspace{-2mm}
\label{fig:bland-altman}
\end{figure}

\subsection{Bland-Altman Plot}
We perform a Bland-Altman analysis on the test set to assess the agreement between the predictions of \model{} and the ground-truth weight measurements (Fig.~\ref{fig:bland-altman}).
The analysis shows negligible bias of \SI{-1.4}{\gram} indicating a strong agreement of the weight estimates against the ground-truths.
Notably the plot largely exhibits homoscedasticity, signifying consistent variability across a significant range of weights.
The 80\% Limits of Agreement (LoA) is $\sim$$\pm$\SI{190}{\gram} which makes the solution acceptable for deployment based on inputs from public health experts.

%% file: tables/data_validation.tex
\begin{table}[t]
\small
\centering
\begin{tabular}{l c c}
\toprule
Criteria & \# discarded visits & \% discarded \\
\midrule
Environment Validation & 53 & 0.3\% \\
Video Quality Validation & 441 & 2.6\% \\
Weight Validation & 3200 & 19.2\% \\
$<$40 frames in video & 17 & 0.1\% \\
\midrule
Total & 3711 & 22.2\% \\
\bottomrule
\end{tabular}
\vspace{-2mm}
\caption{Number of visits discarded based on all the data validation criteria.}
\label{table:data_validation}
\end{table}

%% file: tables/weight_validation.tex
\begin{table}[t]
\small
\centering
\begin{tabular}{l c}
\toprule
Criteria & \# discarded visits \\
\midrule
Newborn is not visible & 20 \\
Newborn is wearing clothes & 47 \\
Readings beyond \SI{50}{\gram} of each other & 982 \\
Other problems & 2151 \\
\midrule
Total & 3200 \\
\bottomrule
\end{tabular}
\vspace{-2mm}
\caption{Number of visits discarded due to failure in one or more weight validation criteria.
We apply rules strictly to ensure high quality and accurate ground-truth, both for training and evaluation.
See Sec.~\ref{subsec:app:gtweight} for a detailed explanation.}
\label{table:weight_validation}
\end{table}

%% file: tables/handcrafted_feature_model_info.tex
\begin{table}[t]
\small
\centering
\captionsetup{skip=2mm, font=small}
\begin{tabular}{l c}
\toprule
Representation & Feature dimensionality \\
\midrule
Hu moments & 350\\
Regionprops & 300\\
HOG & 7200 \\
\bottomrule
\end{tabular}
\caption{Hand-crafted feature dimensionality across 25 frames.}
\label{table:hand_craft_fea_model_info}
\end{table}

%% file: tables/baseline_ablations_extended.tex
\begin{table}[t]
\small
\centering
\captionsetup{skip=2mm, font=small}
\begin{tabular}{l c c c }
\toprule
Representation & Feature Scaling & Model & W (\unit{\gram}) \\
\midrule
HOG & $z$-score & LR & 853.5 \\
HOG &  minmax & MLP & 578.4 \\
HOG &  $z$-score & SVR & 466.7 \\
\midrule
Hu moments &  log & LR & 475.1 \\
Hu moments &  log & MLP & 477.6 \\
Hu moments &  $z$-score & SVR & 470.3 \\
\midrule
Regionprops &  z-score & LR & 401.9 \\
Regionprops &  minmax & MLP & 446.5 \\
Regionprops &  $z$-score & SVR & 399.4 \\
\midrule
Hu + Regionprops &  log + $z$-score & LR & 398.1 \\
Hu + Regionprops &  log + $z$-score & MLP & 529.0 \\
Hu + Regionprops &  $z$-score & SVR & 393.0 \\
\bottomrule
\end{tabular}
\caption{Performance of hand-crafted features on weight estimation with the validation set.}
\label{table:baseline_ablations_extended}
\end{table}